\documentclass[pmlr]{jmlr}
\RequirePackage{graphicx}
 \usepackage{booktabs}
\usepackage{longtable}
\makeatletter
\def\set@curr@file#1{\def\@curr@file{#1}} 
\makeatother
\usepackage[load-configurations=version-1]{siunitx} 
\usepackage{listings}
\definecolor{codegray}{gray}{1}

\lstdefinestyle{mystyle}{
    backgroundcolor=\color{codegray},
    breakatwhitespace=false,         
    breaklines=true,                 
    captionpos=b,                    
    keepspaces=true,                 
    showspaces=false,                
    showstringspaces=false,
    showtabs=false,                  
    tabsize=2,
    frame=single, 
    breakindent=0pt,
    basicstyle=\ttfamily\footnotesize,
}
\jmlrvolume{[VOLUME \# TBD]}
\jmlryear{2024}
\jmlrworkshop{Machine Learning for Healthcare}

\title[Unimodal Approach to Multimodal Challenges in Radiology]{Simplifying Multimodality: Unimodal Approach to Multimodal Challenges in Radiology with General-Domain Large Language Model}

\author{\Name{Seonhee Cho}
       \Email{seonhee.cho@kaist.ac.kr}\\ 
       \addr 
       Kim Jaechul Graduate School of AI\\
       KAIST\\
       Seongnam, Gyeonggi, Korea 
       \AND
       \Name{Choonghan Kim} 
       \Email{choonghankim@kaist.ac.kr} \\
       \addr 
       Moon Soul Graduate School of Future Strategy\\
       KAIST\\
       Daejeon, Korea
       \AND
       \Name{Jiho Lee} \\
       \addr 
       Department of Computer Science and Engineering\\
       Ewha Womans University\\
       Seoul, Korea
       \AND
       \Name{Chetan Chilkunda} \\
       \addr 
       Department of Chemical Engineering\\
       Carnegie Mellon University\\
       Pittsburgh, PA, USA
       \AND
       \Name{Sujin Choi} \\
       \addr 
       Division of Environmental Science and Engineering\\
       POSTECH\\
       Pohang, Gyeongbuk, Korea
       \AND
       \Name{Joo Heung Yoon\nametag{\thanks{Corresponding author}}}
       \Email{yoonjh@upmc.edu}\\ 
       \addr 
       Division of Pulmonary, Allergy, Critical Care, and Sleep Medicine\\
       Department of Medicine\\
       University of Pittsburgh\\
       Pittsburgh, PA, USA}

\begin{document}

\maketitle

\begin{abstract}
  Recent advancements in Large Multimodal Models (LMMs) have attracted interest in their generalization capability with only a few samples in the prompt. This progress is particularly relevant to the medical domain, where the quality and sensitivity of data pose unique challenges for model training and application. However, the dependency on high-quality data for effective in-context learning raises questions about the feasibility of these models when encountering with the inevitable variations and errors inherent in real-world medical data.
  In this paper, we introduce MID-M, a novel framework that leverages the in-context learning capabilities of a general-domain Large Language Model (LLM) to process multimodal data via image descriptions. MID-M achieves a comparable or superior performance to task-specific fine-tuned LMMs and other general-domain ones, without the extensive domain-specific training or pre-training on multimodal data, with significantly fewer parameters. This highlights the potential of leveraging general-domain LLMs for domain-specific tasks and offers a sustainable and cost-effective alternative to traditional LMM developments. Moreover, the robustness of MID-M against data quality issues demonstrates its practical utility in real-world medical domain applications.
  %
  
\end{abstract}

\section{Introduction}

Recently, large multimodal models (LMMs) have made significant progress, becoming capable of interpreting and integrating information across different data types, such as text and images~\citep{achiam2023gpt, driess2023palm}. This progress closely aligns to advancements in Large Language Models (LLMs), since many LMMs are built upon the structural foundation of pre-trained LLMs~\citep{touvron2023llama, chiang2023vicuna}. 
Consequently, attempts to use LMM's in-context learning capability without fine-tuning on specific tasks are increasing~\citep{wei2022chain, zhang2023multimodal}. This tendency also appears in the medical domain, where the data collection is challenging due to the concerns on using sensitive clinical information containing re-identifiable patient data~\citep{benitez2010evaluating} and the expensive labeling process which requires expert knowledge.

However, the performance of in-context learning approach is highly susceptible to the data quality~\citep{liu2023pre}. One primary cause of degrading data quality is data loss occurring during the data collection and curation process, even when guided by expert input.
For instance, the analysis of radiographic images has reported an error rate of 3 to 5\% \citep{robinson1999variation} and the problem appears challenging to resolve until now \citep{brady2017error}. Similarly, a basic chart review showed electronic health record (EHR) errors could reach 9 to 10\% \citep{feng2020transcription}. 
Additionally, variations in data interpretation by medical professionals are quite common~\citep{Whitinge008155} and this can further exacerbate these issues by adding another layer of complexity to ensuring the quality and consistency of data.

In this paper, we introduce \textbf{MID-M}, a \textbf{M}ultimodal framework with \textbf{I}mage \textbf{D}escription for \textbf{M}edical domain. 
It is an in-context learning framework that demonstrates robust performance even with low-quality data. Our framework uses general-domain LLM and leverages multimodal data by converting images to textual descriptions. Notably, it achieves comparable performance to other general-domain and fine-tuned LMMs without pre-training on multimodality and extensive fine-tuning for the medical domain. In addition, by processing the image as a text description, it has the advantage of representing the image in an accessible and interpretable way, compared to traditional embedded vector representations. Our framework is illustrated in Figure 1.

\begin{figure}[h]
\caption{Overview of \textbf{MID-M} framework.}
\centering
\includegraphics[width=\textwidth]{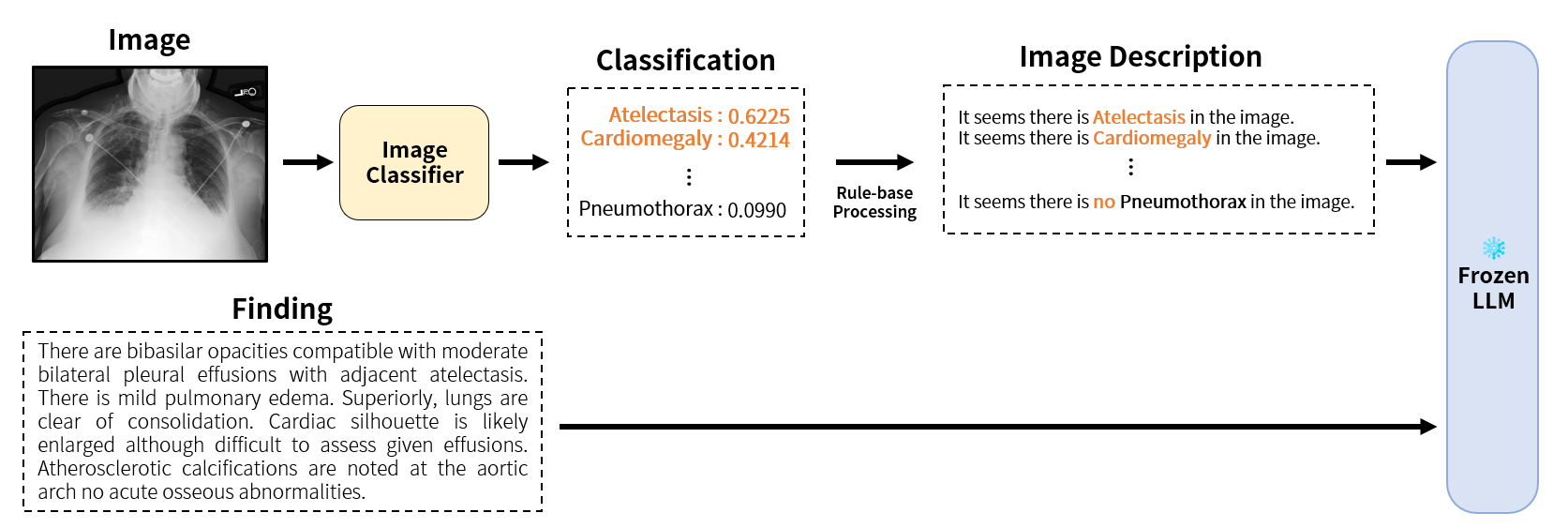}
\end{figure}

To evaluate the model's performance in the scenario with low-quality data, we systematically mask the texts in a medical dataset and make the new adversarial evaluation data.
We compare the performance of general domain pre-trained and medically fine-tuned LMM models to our framework in a few-shot setting, focusing on their accuracy and semantic comprehensibility. Through this experiment, we aim to assess the effectiveness of LMMs in generalizing to domain-specific tasks and managing incomplete source data.

\subsection*{Generalizable Insights about Machine Learning in the Context of Healthcare}

In this study, we analyze the generalization capabilities of LLMs and LMMs across various dimensions.
First, we investigate the ability of models trained in general domains to effectively generalize to medical texts through few-shot learning. This aligns with the recent trend towards adopting eco-friendly approaches in AI model training by reducing the need for massive pretraining and extensive fine-tuning. It also contributes to the validation of model's potential to generalize well across any domain or task, as aimed for in general domain research.
Second, we verify the models' ability to generalize in situations where data quality deteriorates. The challenge that in-context learning models are easily influenced by the number of few-shot samples and the quality of trained data remains unresolved. Therefore, concrete evaluation in this regard is crucial for the advancement of models.

Furthermore, our research reveals that text-only models can efficiently handle multimodal healthcare tasks with less computational demand. We demonstrate that smaller language models can achieve results comparable to larger models, indicating a path for wider adoption of AI in healthcare, especially in resource-limited settings in computation or a lower level of healthcare environment. This goes in line with a move towards developing globally accessible AI solutions by the National Institutes of Health\footnote{https://commonfund.nih.gov/bridge2ai}, highlighting the potential utility of proposed research in expanding AI's reach.

\section{Background}

In this section, we review the relevant literature on large language or multimodal models in the general domain as well as those applied specifically to the tasks in healthcare.

\subsection{Large Language Models for Vision and Language Tasks}
\paragraph{Multimodal pretraining}

Large multimodal models have often leveraged language models as the backbone architecture and incorporate methods to connect the image encoder with the language model~\citep{awadalla2023openflamingo, alayrac2022flamingo, li2023mimic, li2023blip2, xu2024llava, laurenccon2024obelics, liu2024llava, dai2024instructblip}. Notably, LLaVA~\citep{liu2024llava} employs a direct projection layer, BLIP-2~\citep{li2023blip2} uses a lightweight Q-Former, and Flamingo~\citep{alayrac2022flamingo} uses gated cross-attention, to name a few examples.

However, this paradigm suffers from two critical drawbacks. First, pretraining requires a high computational burden. Jointly aligning vision backbones and large language models requires large computational resources, despite considerably fewer trainable parameters compared to the total parameters. For example, Flamingo takes two weeks with 1,536 TPUs for the largest run, and BLIP-2 takes around 10 days with 16 A100 GPUs. Hence, it becomes prohibitively expensive to switch to a different large language model. The second limitation is the lack of modularity and flexibility.
Even though they integrate pretrained image encoder and frozen LLM, updating this integrated system demands training the network with substantial data instances and time~\citep{laurenccon2024obelics}, hindering iterative research and development.

\paragraph{Multimodal Reasoning with Language}
There have also been approaches that attempt to tackle multimodal tasks using language models alone~\citep{yang2022empirical, guo2022images, hu2022promptcap, xie2022visual, suris2023vipergpt, berrios2023lens}.
Among these, LENS~\citep{berrios2023lens} is particularly relevant to our work, as this framework extracts generic information such as tags, attributes, and captions with three different vision modules, and provides them to the language model. While this method can be beneficial for general domain tasks, such an approach does not scale well to tasks that require expert knowledge. In contrast, our framework facilitates LLM's multimodal reasoning within the medical domain. We introduce a more streamlined framework with a single `vision module' and tailor a general domain LLM specifically for medical tasks, by leveraging the adaptability of the LLM in this knowledge-intensive domain.

\subsection{Large Language Models for Healthcare Domain}
Several efforts have been made to specialize these general-purpose foundation models through fine-tuning on healthcare domain-specific data~\citep{li2024llava, moor2023med, zhang2023biomedgpt, wutowards, tu2024towards, sun2023pathasst}. However, task-specific fine-tuning is often inefficient, requiring large amounts of high-quality medical data and extensive computing resources.

Consequently, efficient adaptation methods are also explored, which primarily follow two categories: lightweight fine-tuning techniques~\citep{hu2021lora, houlsby2019parameter}, and leveraging the models' in-context learning capabilities in a few-shot setting. While fine-tuned models generally outperform their counterparts in few-shot setting\citep{van2023clinical, van2023radadapt, lewis-etal-2020-pretrained}, there is a growing interest in exploring domain adaptation techniques, mainly in-context learning, which do not require additional training~\citep{agrawal2022large, lievin2023can, byra2023few, yan2023robust, nori2023can, van2023radadapt}. 
In-context learning has many potential advantages for the clinical domain because there is often a limited set of labeled data due to the high level of expertise needed for annotation.
Our work also aligns with this in-context learning approach.
By leveraging the capabilities of general-purpose LLMs without additional fine-tuning, our approach aims to overcome the limitations of task-specific fine-tuning while maintaining the models' broad knowledge and generalization abilities.

\section{Methods}
\label{sec:method}
This section introduces \textbf{MID-M}, an efficient in-context learning framework that performs multimodal tasks using only small-sized language models. MID-M consists of two key components: a domain-specific image converter and a general-domain language model.

\paragraph{Image Conversion with Domain-Specific Image Classifier}

To reformulate vision-language tasks to text-to-text task, we first extract key information from images and convert it to text. Taking inspiration from \textit{Classification-via-Description}~\citep{menon2022visual} which showed good generalization performance, we employ a \textit{Description-via-Classification} strategy. \textit{Classification-via-description} involves generating textual features representing each class with a language model, then analyzing whether the input exhibits those features to make the final classification decision. Conversely, we use the classification result to generate an image description.
Note that using a general-domain image module trained only on generic data, as in \cite{berrios2023lens}, cannot yield meaningful classification results since chest X-rays require domain knowledge. 
Therefore, we utilize a SOTA model~\citep{chong2023category} for CheXpert, a publicly available large chest X-ray classification dataset. This classifer computes a probability of 14 major diseases for chest radiology images. For diseases that the classifier predicts to have a probability of presence greater than 20\%, we use the sentence “It seems there is \{\textit{disease}\} in the image.” Conversely, for diseases predicted as absent, we use “It seems there is no \{\textit{disease}\} in the image.” These sentences describing the presence or absence of each disease are all concatenated and used as an image description. This method can be extended to other domains without domain expertise, by simply replacing the classifier with a domain-specific one. We provide an example of our prompt in Appendix A.

\paragraph{General-domain LLM}
We utilize a general-domain language model for multimodal reasoning. While chains-of-thought~\citep{wei2022chain} are generally advantageous for domain adaptation and complex reasoning, their benefits are not consistent across generation tasks~\citep{kim2023cot}. Furthermore, it requires domain knowledge to generate rationales. Thus, we utilize strategies known to be helpful for domain adaptation~\citep{van2023radadapt} instead. 
Specifically, we prepend a sentence assigning the role of an AI assistant as a domain expert: ``You are an expert medical professional." This is followed by a detailed task description crafted for the radiology domain.
For the few-shot examples, we provide the image description, corresponding finding, and impression in sequence. During inference, we omit the impression from the examples to be completed by the model. We use a BM25 retriever, known for its capability to compute token overlap quickly and retrieve relevant examples effectively.

\section{Dataset}
\label{sec:dataset}
We utilized the MIMIC-CXR dataset~\citep{johnson2019mimic}, a de-identified and publicly available chest radiograph database collected from 2011 to 2016. It has a total number of 227,835 images and 128,032 paired reports from 65,379 patients. Paired reports include \textit{finding} and \textit{impression}. A \textit{finding} is a detailed observation made from the X-ray images written by radiologists, and an \textit{impression} is a concise summary of the \textit{finding} which becomes the primary mode of communication between medical professionals.

\paragraph{Preliminary Analysis}
Our examination of the MIMIC-CXR dataset revealed 3,146 samples (2.46\% of the total) that were unsuitable for research due to one or more of the following reasons: the \textit{finding} label contained fewer than three words, was shorter than the \textit{impression} label, or had more than three words masked. This level of incompleteness aligns with error rates reported in previous studies~\citep{robinson1999variation}.

\paragraph{Pre-preprocessing}
To prepare the data for our study, we first merged the original training, validation, and test sets. We then removed samples with the \textit{findings} at both extremes of length, removing the shortest and longest 25\% of findings to ensure a meaningful level of detail remained after applying masking techniques. This process resulted in a dataset of 64,613 images and reports, which we then randomly split into training, validation, and test sets containing 62,613, 1,000, and 1,000 samples, respectively.
\paragraph{Masking}
To further evaluate the robustness of LLMs against incomplete data, we employed a masking technique akin to that used by BERT~\citep{devlin2018bert}, randomly deleting words or phrases at subword levels. We conducted masking at three different rates to the original test set and yielded four distinct test sets: original (\textit{full}), corrupted at 10\% (\textit{corrupted 0.1}), corrupted at 30\% (\textit{corrupted 0.3}), and corrupted at 50\% (\textit{corrupted 0.5}).

We evaluated the effectiveness of our corruption strategy using the F1CheXbert~\citep{zhang2020chexbert} metric.
CheXbert~\citep{smit2020chexbert} is designed to identify 14 medical observations in chest X-ray images. The F1CheXbert score is calculated by comparing CheXbert's predictions on the generated text against the corresponding reference text. We compare the findings from \textit{full} against the findings from each corrupted test set. We discovered a strong correlation between the extent of text corruption and the precision of disease mentions. The micro-average F1CheXbert scores were  91.2\% for the 10\% corrupted set, 69.3\% for the 30\% corrupted set, and 51.1\% for the 50\% corrupted set.
Examples of corrupted text are illustrated in Table \ref{tab:data_example}.

\begin{table}[t]
  \centering 
  \small
  \caption{Example of MIMIC-CXR findings with different masking rate.}
  \begin{tabular}{p{0.21\linewidth}|p{0.21\linewidth}|p{0.21\linewidth}|p{0.21\linewidth}}
  \toprule
    \textbf{Full} & \textbf{Corrupted 0.1} & \textbf{Corrupted 0.3} & \textbf{Corrupted 0.5}\\
    \midrule
    PA and lateral views of the chest provided. AICD projects over the left chest wall with lead tip extending to the region of the right ventricle. The heart is mildly enlarged. There is no evidence of pneumonia or CHF. No effusion or pneumothorax seen. Bony structures are intact. &
    PA and lateral views \_ the chest provided. AICD projects over the \_ chest wall with lead tip extending to the region of the right ventric\_. \_ heart is mildly enlarged. there is no \_ of pneumonia or CHF. \_ effusion or pneumothorax seen. Bony structures are intact. &
    PA and lateral \_ of \_ chest provided. \_CD projects over the left \_ wall with lead tip \_ to the region of the \_ ventric \_ the heart is \_. \_ of pneumonia or CH\_. no effusion or pneum\_rax seen. \_ intact.
    &
    PA and lateral views of \_ provided. AI \_ the \_ chest \_ with \_ tip \_ to the \_ vent \_ heart is \_ enlarged. there \_ no \_ F \_ no \_ usion or pneum \_rax \_ structures are intact \_ \\
    \bottomrule
  \end{tabular}
  \label{tab:data_example} 
\end{table}

\section{Experimental Setup}
\subsection{Task}
In this study, the task requires the model to generate an impression based on an X-ray image and its corresponding findings. For all experiments, we employed a 2-shot learning approach, utilizing the BM25 algorithm to retrieve two training samples with findings similar to the findings of the test sample. When experimenting with corrupted test sets, the samples are retrieved based on the corrupted findings. This task is designed to comprehensively evaluate the model's ability to understand medical images and clinical data, as well as to identify salient information even when incomplete data is given.

\subsection{Baselines}
\label{sec:setup-baseline}
We have two groups of baselines in our experiments: LMMs that are pre-trained in the general domain and those further fine-tuned with medical domain data. 
However, considering the constraints of a few-shot learning setting, we found a limited number of LMMs are capable of processing multiple independent images in a single prompt. Notably, Flamingo~\citep{alayrac2022flamingo}, is trained on interleaved text and image pairs and supports multiple images. We used its open-sourced version, OpenFlamingo~\citep{awadalla2023openflamingo}. This functionality is also applied to the models that are based on Flamingo, IDEFICS~\citep{laurenccon2024obelics} and OTTER~\citep{li2023mimic}.
For models trained on data from the medical domain, we selected MedFlamingo~\citep{moor2023med} and RadFM~\citep{wu2023towards}. 
MedFlamingo is trained with medical publications and textbooks and is based on OpenFlamingo. RadFM is a specialized model for the radiology domain and pre-trained with large-scale, high-quality multimodal dataset named MedMD. It's important to note that although MedMD includes MIMIC-CXR, we still chose to include it in our baseline set. Despite data leakage concerns in our \textit{full} test setting, we believe that evaluating the model's performance on corrupted versions of MIMIC-CXR would provide valuable insights into its robustness and generalization capabilities.


A detailed comparison is presented in Table \ref{tab:baselines}. Note that all the baseline models have 9 billion parameters or more, whereas our backbone model, Flan-T5-xl, has only 3 billion parameters.
For the baselines, we adhered to the example code in their public GitHub repositories when available; otherwise, we utilized a consistent prompt with the example code from other baselines.

\begin{table}[t]
  \centering 
  \caption{Descriptions of baseline models. The first four models are general-domain LMMs that can handle multiple images in one input. The two models in the middle are LMMs that are fine-tuned with medical domain data.}
  \begin{tabular}{l|cccc}
  \toprule
    \textbf{Model} & \textbf{Size} & \textbf{Domain} & \textbf{Image Encoder} & \textbf{Language Model} \\
    \midrule
    Otter$_{\text{M}}$      & 9B   & General   & ViT-L/14    & MPT-7b  \\ 
    Otter$_{\text{L}}$    & 9B  & General   & ViT-L/14    & LLaMA-7b \\ 
    OpenFlamingo        & 9B   & General & ViT-L/14    & MPT-7b     \\ 
    IDEFICS             & 9B   & General & ViT-H/14    & LLaMA-7b   \\ 
    \midrule
    MedFlamingo         & 9B  & Medical  & ViT-L/14    & LLaMA-7b       \\ 
    RadFM               & 14B & Radiology   & ViT-B/32    & MedLLaMA-13b   \\ 
    \midrule
    \textbf{MID-M}         & \textbf{3B}  & \textbf{General}  & \textbf{Classifier(7M)}    & \textbf{Flan-T5-xl}  \\ 
    \bottomrule
  \end{tabular}
  \label{tab:baselines} 
\end{table}

\subsection{Evaluation}

The performance of the models are compared using two different LLM evaluation methods, ROUGE-L~\citep{lin2004rouge} and F1RadGraph~\citep{delbrouck2022improving}.
ROUGE, the Recall-Oriented Understudy for Gisting Evaluation, is a metric widely used in the text summarization task. This score evaluates the similarity between the generated text and the reference text based on token overlap. Specifically, we used ROUGE-L which measures the longest common subsequence (LCS) to assess similarity at the sentence level. 
F1RadGraph is a metric that evaluates medical entities and relations within the generated reports. It utilizes a PubMedBERT \citep{gu2021domain} fine-tuned on the RadGraph \citep{jain2021radgraph}, a dataset with chest X-ray radiology reports annotated with medical entities and their relations. This fine-tuned PubMedBERT model identifies the entities and relations from the generated reports and the reference reports.  When the generated report correctly matches the entities and their relation to that in the reference, they receive a reward of 1, and otherwise, 0. The final F1 score is computed based on this reward score. 
Both metrics are implemented using the Python Package Index (PyPI)\footnote{https://pypi.org/project/radgraph/}


\section{Experiment}
\subsection{Result}
\label{sec:experiment_result}
\begin{table}[t]
  \centering 
  \caption{Model performances in each test set with regard to the accuracy. The highest score is bolded and the second highest is underlined.}
  \begin{tabular}{l|c|c|c|c}
  \toprule
    \textbf{Method} & \textbf{\textit{Full}} & \textbf{\textit{Corrupted 0.1}} & \textbf{\textit{Corrupted 0.3}} & \textbf{\textit{Corrupted 0.5}} \\
    \midrule
    \multicolumn{5}{c}{\textit{ROUGE}} \\
    \midrule
    Otter$_{\text{M}}$      & 0.3140    & 0.2258    & 0.2116  & 0.1894 \\ 
    Otter$_{\text{L}}$    & 0.2655    & 0.2089    & 0.1876 & 0.1461 \\ 
    OpenFlamingo        & \underline{0.4310}    & 0.3458    & \underline{0.3420}    & \underline{0.2888} \\ 
    IDEFICS             & 0.3951    & \underline{0.3776}    & 0.3321    & 0.2807 \\ 
    MedFlamingo         &  0.3804 & 0.3622  & 0.3270  & 0.2784 \\ 
    RadFM               & \textbf{0.5222}    & \textbf{0.3952}    & 0.3299    & 0.2534 \\ 
    \textbf{MID-M}(ours)               & 0.4064   & 0.3719    & \textbf{0.3438}    & \textbf{0.2977} \\ 
    \midrule
    \multicolumn{5}{c}{\textit{F1RadGraph}} \\
    \midrule
    Otter$_{\text{M}}$      & 0.2801    & 0.2114    & 0.1704  & 0.1327 \\ 
    Otter$_{\text{L}}$    & 0.2587    & 0.1994    & 0.1720 & 0.1237 \\ 
    OpenFlamingo        & 0.4030    & 0.3427    & \textbf{0.3151}    & {0.2541} \\ 
    IDEFICS             & \underline{0.4089}    & \underline{0.3688}  & {0.3064}    & \underline{0.2543} \\ 
    MedFlamingo         &  0.3973 & 0.3510 & 0.3046 & 0.2566 \\ 
    RadFM               & \textbf{0.5054}    & \textbf{0.3839}    & \underline{0.3094}    & 0.2363 \\ 
    \textbf{MID-M}(ours)               & 0.3732    & 0.3379    & 0.2973    & \textbf{0.2622} \\     
    \bottomrule
  \end{tabular}
  \label{tab:main_experiment} 
\end{table}

The experimental results are presented in Table \ref{tab:main_experiment}. 
MID-M achieves performance comparable to that of other baseline models in all settings, despite using only a third of the parameters. It even surpasses other models in experiments with corrupted data, highlighting its exceptional generalization capabilities.
RadFM, which previously trained on MIMIC-CXR, achieves the highest score on the \textit{full} test set, which is not a surprising result. IDEFICS and OpenFlamingo demonstrate strong generalization ability, achieving scores around 0.4 across both ROUGE and F1RadGraph metrics. 

However, we observed that some models are highly sensitive to even a minor level of data corruption (i.e. 0.1). For instance, RadFM suffers from a significant performance drop in corrupted settings and shows performance similar to or even worse than large-scale general domain models. This sensitivity to text corruption is likely due to its training on only high-quality radiology images and associated texts. It is critical problem in clinical applications where even small decreases in performance can significantly impact patient care. Otter and OpenFlamingo also shows considerable performance declines in corrupted settings. Conversely, IDEFICS, MedFlamingo and our framework maintain robust performance on low-quality data.
As the text masking probability increases (i.e. 0.3, 0.5), our model begins to demonstrate its full potential. It achieves the highest scores under the ROUGE metric and exhibits either comparable or superior performance on the F1RadGraph metric.

It is also surprising that MedFlamingo, which is further fine-tuned from OpenFlamingo with medical domain text, does not outperform OpenFlamingo. This may be because MedFlamingo is primarily fine-tuned for medical visual question-answering tasks rather than various medical generation tasks. These experimental results shows the importance of task setting during pre-training and fine-tuning.

\subsection{Ablation Studies}
To comprehend the the effect of each component in our framework to the overall performance, we systematically remove each element - textual input (findings), image input (descriptions), and both. 
In scenarios where both text and images are excluded, the model must generate impressions based solely on two impressions retrieved based on the similarity of findings. The results of these ablation studies are presented in Table \ref{tab:ablation}. 

The results indicate that the multimodal framework exhibits the strongest generalization capability by leveraging all available information. 
Since the impressions are primarily derived from findings, the text component alone proves to be most effective when the text data encompasses sufficient information. Still, using both image and text information achieves comparable scores and outperforms in the corrupted test sets.
When text corruption exceeds a 10\% probability, incorporating image descriptions improves the model's accuracy compared to using only the findings or only the image descriptions in most cases. One interesting point is, at the highest level of text corruption, the corrupted text acts as noise and severely impairs the model's performance, leading to worse outcomes than when both text and images were excluded.
This trend remains consistent across evaluation metrics of token overlap, measured by ROUGE, and semantic coherence, measured by F1RadGraph.

\begin{table}[t]
  \small 
  \centering 
  \caption{Model performances in different input settings.}
  \begin{tabular}{l|c|c|c|c}
  \toprule
    \textbf{Method} & \textbf{\textit{Full}} & \textbf{\textit{Corrupted 0.1}} & \textbf{\textit{Corrupted 0.3}} & \textbf{\textit{Corrupted 0.5}} \\
    \midrule
  \multicolumn{5}{c}{\textit{ROUGE}} \\
    \midrule
    \textbf{MID-M}               & \underline{0.4064}   & \textbf{0.3719}    & \underline{0.3438}    & \textbf{0.2977} \\ 
    w/o text              & 0.3794   & 0.3563    & \textbf{0.3440}    & \underline{0.2942} \\ 
    w/o image             & \textbf{0.4119}  & \underline{0.3683}    & 0.3218    & 0.2550 \\ 
    w/o text and image    & 0.3649  & 0.3309    & 0.3129   & 0.2670 \\ 

    \midrule
    \multicolumn{5}{c}{\textit{F1RadGraph}} \\
    \midrule
    \textbf{\textbf{MID-M}}               & \underline{0.3732}    &\textbf{ 0.3379}    & \textbf{0.2973}    & \textbf{0.2622} \\     
    w/o text    & 0.3354    & 0.3185    & \underline{0.2948} & \underline{0.2575} \\ 
    w/o image         & \textbf{0.3918}    & \underline{0.3323}    & 0.2784    & 0.2247 \\ 
    w/o text and image         & 0.3298    & 0.2976    & 0.2783    & 0.2378 \\ 
    \bottomrule
  \end{tabular}
  \label{tab:ablation} 
\end{table}



\subsection{Disease Identification Ability}

In this section, we further analyze the model's ability to identify the major diseases in X-rays and corresponding findings. The evaluation in Section \ref{sec:experiment_result} is based on a broad comparison of impressions and predictions at the word and semantic level. Additionally in this section, we use the F1CheXbert to assess whether the model identifies all the key diseases noted in the gold impressions in its predictions as well. We report both the micro-average F1 score and the individual F1 scores for five main diseases, as shown in Table \ref{tab:chexbert}.

\begin{table}[h]
\centering
\small
\caption{Comparison of MID-M's performance on the identification of major diseases under different levels of data corruption. Abbreviations: CMG, cardiomegaly; Consol, consolidation; Atelect, atelectasis; PE, pleural effusion.}

\label{tab:chexbert}

\begin{tabular}{@{} c|c|c|c|c|c|c @{}}
\toprule
& \textbf{CMG} & \textbf{Edema} & \textbf{Consol} & \textbf{Atelect} & \textbf{PE} & {Micro Avg.} \\
\midrule
\textbf{\textit{Full}} & 0.4200 & 0.5714 & 0.4444 & 0.4916 & 0.6335 & 0.5302 \\
\textbf{\textit{Corrupted 0.1}} & 0.3204 & 0.4923 & 0.1538 & 0.4067 & 0.5543 & 0.4379 \\
\textbf{\textit{Corrupted 0.3}} & 0.3010 & 0.3884 & 0.1176 & 0.3972 & 0.4250 & 0.3724 \\
\textbf{\textit{Corrupted 0.5}} & 0.2631 & 0.3606 & 0.1538 & 0.2704 & 0.4467 & 0.3327 \\
\midrule
{Support} & 87 & 148 & 25 & 151 & 138 & 549 \\
\bottomrule
\end{tabular}
\end{table}

We found that the model's performance declines with increased data corruption, but the rate of decline varies across diseases. For instance, the model's ability to identify pleural effusion (PE) remains relatively robust against corruption, even under conditions of high data corruption, compared to other diseases. In contrast, our model experiences a significant drop in consolidation (Consol) even at the lowest level of corruption (0.1). This suggests that the model manages to recognize some diseases in a manner less susceptible to corruption.
This might be attributed to the feasibility of retrieving similar samples. When there are many samples with a certain disease in the training set, it becomes easier to retrieve relevant samples for the diseases and vice versa.
In addition, a significant drop in consolidation might be attributed to a unique linguistic challenge. When some of the subwords from consolidation are obscured, the remaining segments (such as '\textit{solid}' or '\textit{-tion}') could be mistakenly associated with other terms frequently used in radiology reports, complicating its identification.

\subsection{In-Context Learning Ability}
In this section, we evaluate the model's in-context learning capabilities by varying the number of examples (shots) included in the prompt. Previous experiments, as discussed in Section \ref{sec:experiment_result}, uniformly used a two-shot setting. In addition to this, we explore the model's abilities in zero-shot and one-shot settings. The results are presented in Table \ref{tab:in-context}.

\begin{table}[t]
  \small 
  \centering 
  \caption{Model performances in different number of shots in the input prompt. }
  \begin{tabular}{l|c|c|c|c}
  \toprule
    \textbf{Method} & \textbf{\textit{Full}} & \textbf{\textit{Corrupted 0.1}} & \textbf{\textit{Corrupted 0.3}} & \textbf{\textit{Corrupted 0.5}} \\
    \midrule
    \multicolumn{5}{c}{\textit{ROUGE}} \\
    \midrule
    0-shot             & 0.1673  & 0.1270    & 0.1127    & 0.1091 \\ 
    1-shot              & 0.3818   & 0.3438    & {0.3172}    & {0.2180} \\ 
OURS (2-shot)  & 0.4064   & 0.3719    & {0.3438}    & {0.2977} \\    
    \midrule
    \multicolumn{5}{c}{\textit{F1RadGraph}} \\
    \midrule
    
    0-shot             & 0.1573  & 0.0994    & 0.072    & 0.0439 \\ 
    1-shot              & 0.3536   & 0.3091    & 0.2797    & 0.2386 \\ 
OURS (2-shot)  & 0.3732    & 0.3379    & 0.2973    & {0.2622} \\  

    \bottomrule
  \end{tabular}
  \label{tab:in-context} 
\end{table}

We observe a decline in performance in both zero-shot and one-shot settings compared to the two-shot scenario. However, the disparity in performance across these settings is noteworthy. In the zero-shot setting, the model struggles significantly, achieving a ROUGE score of only about 0.1673 on the full test set and with the F1RadGraph score approaching zero across all corrupted test sets. In contrast, introducing just one relevant example enhances the model's performance substantially, achieving results that are comparatively close to those in the two-shot setting in many cases. This underscores the impressive in-context learning capabilities of pretrained large models, highlighting their potential for easy adaptation to different domains without the need for fine-tuning.

\section{Discussion} 
In this paper, we introduce MID-M, a novel multimodal framework designed for the medical domain, which leverages the in-context learning capabilities of a general-domain LLM. By transforming images into textual descriptions, MID-M facilitates an interpretable representation of medical images, enabling complex information to be more accessible to engineers and, potentially, to practitioners.
Notably, MID-M demonstrates superior generalization capability, outperforming models with substantially more parameters that are extensively trained for multimodality and medical applications. This emphasizes the potential of leveraging general-domain models for specialized tasks, offering a sustainable and cost-effective approach without massive pretraining and extensive fine-tuning. This approach aligns with the ongoing shift towards developing powerful AI tools in healthcare.
Furthermore, it highlights the importance of robustness and generalization in systems, particularly in healthcare settings where data quality can vary significantly. The framework's ability to maintain robust performance even with degraded data quality, presents a compelling case for its application in real-world medical scenarios even with limited resources. We hope our work can contribute to the development of globally accessible AI systems.

\paragraph{Limitations}
Despite the contributions of our research, there are a few limitations to our work.
First, while MID-M demonstrates superior in-context learning capability, it still underperforms compared to task-specific fine-tuned models. This is observed from the performance of RadFM on the \textit{full} test set. However, it is important to note that models like RadFM demand high computational resources for their training and show a significant drop in generalization ability when faced with incomplete data. In future works, we could focus on developing in-context learning approaches that can achieve performance comparable to that of task-specific models, even under ideal conditions.
The second limitation is the artificial nature of our masking approach. In real-world applications, masking is likely to occur at the word level and targets sensitive information. Our method, however, employed subword-level random masking.  Despite this fundamental difference, we believe our test set includes samples with realistically plausible masking since sub-word masking cumulatively approximates word-level masking, and sensitive information can also be obscured by randomly chosen. Our statistical analysis, which shows main diseases are masked in proportion to the corruption rate, also supports our hypothesis. Moreover, this approach allows for the systematic masking with which we produced four different test sets. We believe the insights derived from our experiments can provide valuable guidance for future research.

\acks{
This work was supported by Institute of Information \& communications Technology Planning \& Evaluation (IITP) grant funded by the Korea
government(MSIT) (RS-2022-00143911,AI Excellence Global Innovative Leader Education Program). Yoon JH has received grant support from NIH K23GM138984.
}

\bibliography{final}

\newpage
\appendix
\section*{Appendix A. Example of Prompt}
\label{app:a}
Our prompt is composed of the role assignment, task description, two retrieved examples, and the test sample.
Here we provide one example of the prompt that we used for our framework.

\footnotesize 
\begin{lstlisting}[breaklines=true]
You are an expert medical professional. Write a concise summary of the following chest X-ray report. The text description of the X-ray images and the full report, named "Finding" will be provided. Focus on the key findings and diagnoses noted primarily in the image description, while also incorporating relevant details from the full report. The summary, named "Impression", should be concise and presented in correct English sentences.

Image description: There is Atelectasis in the image in 24.20 probability.
There is Cardiomegaly in the image in 1.82 probability.
There is Consolidation in the image in 3.99 probability.
There is Edema in the image in 0.80 probability.
There is Enlarged Cardiomediastinum in the image in 8.94 probability.
There is Fracture in the image in 10.79 probability.
There is Lung Lesion in the image in 7.47 probability.
There is Lung Opacity in the image in 18.30 probability.
There is No Finding in the image in 31.68 probability.
There is Pleural Effusion in the image in 3.73 probability.
There is Pleural Other in the image in 1.05 probability.
There is Pneumonia in the image in 4.88 probability.
There is Pneumothorax in the image in 6.48 probability.
There is Support Devices in the image in 10.13 probability.
Finding: The heart size is normal. The hilar and mediastinal contours are unremarkable. The lungs are well expanded and clear. There is no evidence of a pneumothorax or pleural effusion. The visualized osseous structures are unremarkable.
Impression: No acute cardiopulmonary processes. Specifically, no evidence of an infiltrative process suggestive of pneumonia.

Image description: There is Atelectasis in the image in 2.38 probability.
There is Cardiomegaly in the image in 0.39 probability.
There is Consolidation in the image in 0.76 probability.
There is Edema in the image in 0.10 probability.
There is Enlarged Cardiomediastinum in the image in 3.32 probability.
There is Fracture in the image in 7.94 probability.
There is Lung Lesion in the image in 1.50 probability.
There is Lung Opacity in the image in 4.21 probability.
There is No Finding in the image in 78.47 probability.
There is Pleural Effusion in the image in 0.62 probability.
There is Pleural Other in the image in 0.93 probability.
There is Pneumonia in the image in 0.78 probability.
There is Pneumothorax in the image in 1.96 probability.
There is Support Devices in the image in 9.13 probability.
Finding: The heart size is normal. The hilar and mediastinal contours are unremarkable. The lungs are well expanded and clear. There is no evidence of a pneumothorax or pleural effusion. The visualized osseous structures are unremarkable.
Impression: Normal chest x-ray. Specifically, no pulmonary evidence of TB.

Image description: There is Atelectasis in the image in 4.65 probability.
There is Cardiomegaly in the image in 0.89 probability.
There is Consolidation in the image in 1.28 probability.
There is Edema in the image in 0.21 probability.
There is Enlarged Cardiomediastinum in the image in 4.43 probability.
There is Fracture in the image in 14.87 probability.
There is Lung Lesion in the image in 4.14 probability.
There is Lung Opacity in the image in 5.80 probability.
There is No Finding in the image in 45.93 probability.
There is Pleural Effusion in the image in 3.02 probability.
There is Pleural Other in the image in 3.44 probability.
There is Pneumonia in the image in 1.32 probability.
There is Pneumothorax in the image in 2.53 probability.
There is Support Devices in the image in 5.05 probability.
Finding: The heart size is normal. The hilar and mediastinal contours are unremarkable. The lungs are slightly hyperinflated, however appear to be clear. There is no evidence of pneumothorax or pleural effusions. The visualized osseous structures are unremarkable.
Impression:
\end{lstlisting}
\normalsize 


\end{document}